%% file: 0-main.tex
\documentclass[10pt,twocolumn,letterpaper]{article}

\usepackage{cvpr}              

\usepackage{graphicx}
\usepackage{amsmath}
\usepackage{amssymb}
\usepackage{booktabs}
\usepackage{pifont}

\usepackage{bm}
\usepackage{enumitem}
\usepackage{makecell}
\usepackage{multirow}
\usepackage{soul}
\usepackage{tabularx}
\usepackage{xcolor}

\newcommand{\best}[1]{{\color{blue}{\textbf{#1}}}} 
\newcommand{\second}[1]{{\color{black}{\textbf{#1}}}}

\newcommand{\ours}{VILA}
\newcommand{\vect}[1]{\boldsymbol{#1}}

\newcommand{\cmark}{\ding{51}}%
\usepackage[pagebackref,breaklinks,colorlinks]{hyperref}

\usepackage[capitalize]{cleveref}
\crefname{section}{Sec.}{Secs.}
\Crefname{section}{Section}{Sections}
\Crefname{table}{Table}{Tables}
\crefname{table}{Tab.}{Tabs.}

\def\cvprPaperID{2077} 
\def\confName{CVPR}
\def\confYear{2023}

\begin{document}

\title{\ours: Learning Image Aesthetics from User Comments \\ with Vision-Language Pretraining}

\author{
Junjie Ke, Keren Ye, Jiahui Yu, Yonghui Wu, Peyman Milanfar, Feng Yang \\
Google Research\\
{\tt\small  {\{junjiek, yek, jiahuiyu, yonghui, milanfar, fengyang\}@google.com}}
}

\maketitle

\begin{abstract}
Assessing the aesthetics of an image is challenging, as it is influenced by multiple factors including composition, color, style, and high-level semantics. Existing image aesthetic assessment (IAA) methods primarily rely on human-labeled rating scores, which oversimplify the visual aesthetic information that humans perceive. Conversely, user comments offer more comprehensive information and are a more natural way to express human opinions and preferences regarding image aesthetics. In light of this, we propose learning image aesthetics from user comments, and exploring vision-language pretraining methods to learn multimodal aesthetic representations. Specifically, we pretrain an image-text encoder-decoder model with image-comment pairs, using contrastive and generative objectives to learn rich and generic aesthetic semantics without human labels.  To efficiently adapt the pretrained model for downstream IAA tasks, we further propose a lightweight rank-based adapter that employs text as an anchor to learn the aesthetic ranking concept. Our results show that our pretrained aesthetic vision-language model outperforms prior works on image aesthetic captioning over the AVA-Captions dataset, and it has powerful zero-shot capability for aesthetic tasks such as zero-shot style classification and zero-shot IAA, surpassing many supervised baselines. With only minimal finetuning parameters using the proposed adapter module, our model achieves state-of-the-art IAA performance over the AVA dataset. \footnote{Our model is available at  \url{https://github.com/google-research/google-research/tree/master/vila}}
\end{abstract}

\input{1-introduction}
\input{2-related-work}
\input{3-method}
\input{4-experiment}

\input{5-conclusion}
\clearpage
{\small
\bibliographystyle{ieee_fullname}
\bibliography{egbib}
}

\clearpage
\input{6-supplementary}

\end{document}

%% file: 1-introduction.tex
\section{Introduction}
\label{sec:intro}

\begin{figure}[t]
    \centering
    \includegraphics[width=1\linewidth]{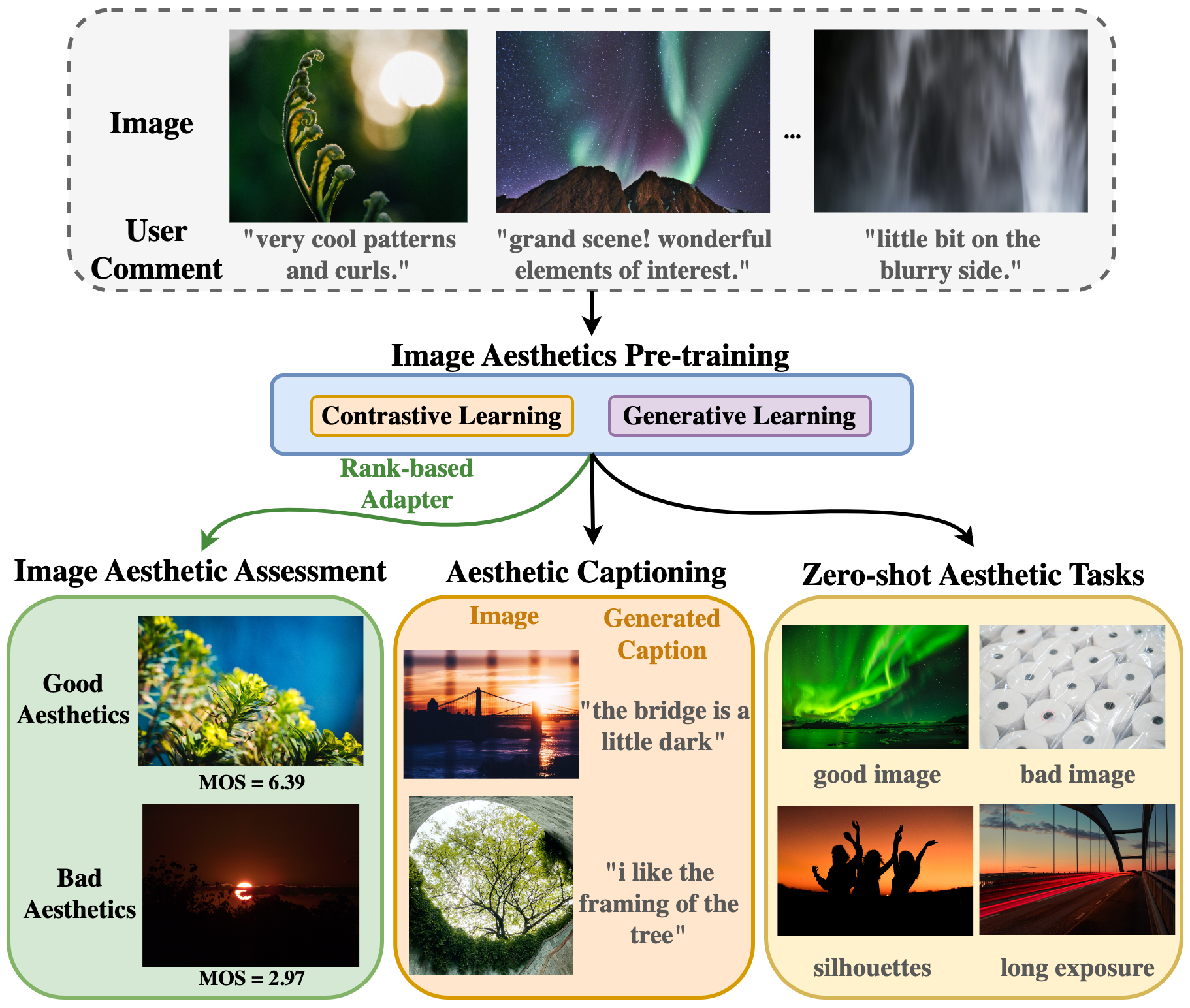}\vspace{-0.5mm}
    \caption{We present \ours, a vision-language aesthetics learning framework based on image and user comment pairs. By pretraining on a contrastive and generative target, it shows superior performance on aesthetic captioning as well as zero-shot aesthetic tasks, \eg, IAA, and style classification. With a lightweight rank-based adapter, we can efficiently adapt the pretrained model to IAA.  }\vspace{-2.5mm}
    \label{fig:teaser} 
\end{figure}

\noindent Image Aesthetic Assessment (IAA) aims to quantify the human perceived aesthetics of an image. It has many important applications, including photo recommendation, selection, and editing. IAA is challenging because it is inherently subjective, and depends on various factors including image composition, color usage, photographic style, and subject matter. In recent years, various learning-based IAA methods have been proposed by leveraging deep models such as convolutional neural networks (CNN)~\cite{talebi2018nima, Chen_2020_CVPR, he_2022_ijcai, Hosu_2019_CVPR} and transformers~\cite{Ke_2021_ICCV}. These approaches learn from human-labeled IAA datasets where images are paired with aesthetic ratings, and models are trained to regress towards the mean opinion scores (MOS).

Directly learning IAA models on human-labeled aesthetic ratings, such as MOS, can be suboptimal as it lacks context regarding why an image is aesthetically pleasing or not. To provide richer supervision, various methods have attempted to integrate external knowledge such as theme~\cite{he_2022_ijcai,niu2022comment}, human eye fixation~\cite{ghosal2019geometry}, and aesthetic attributes~\cite{dhar2011high,kong2016photo}, to enhance IAA performance. These approaches typically rely on multitask training or cascade score prediction with a frozen attribute network. However, obtaining additional labeled data or off-the-shelf models for such methods can be costly.

Compared to the aforementioned methods that require additional annotations, our approach utilizes the abundance of image-comment pairs available on aesthetic websites and photographic forums.  These pairs can be easily obtained from the Internet and contain extensive aesthetic information (\eg objects, themes, styles, and user emotions), since humans are better at expressing aesthetic preferences  through natural language than through abstract scores.
On image sharing platforms like Flickr and DPChallenge\footnote{\url{https://www.dpchallenge.com/}}, user comments offer valuable insights into how they evaluate an image's aesthetics. For instance, as shown in Fig.~\ref{fig:teaser} (top), comments such as ``very cool patterns and curls" and ``little bit on the blurry side" reflects users' positive and negative aesthetic opinions respectively. We aim to learn the diverse aesthetic semantics present in these image-comment pairs to establish a solid foundation for downstream IAA tasks. 

\begin{figure*}[t]
    \centering
    \includegraphics[width=1\linewidth]{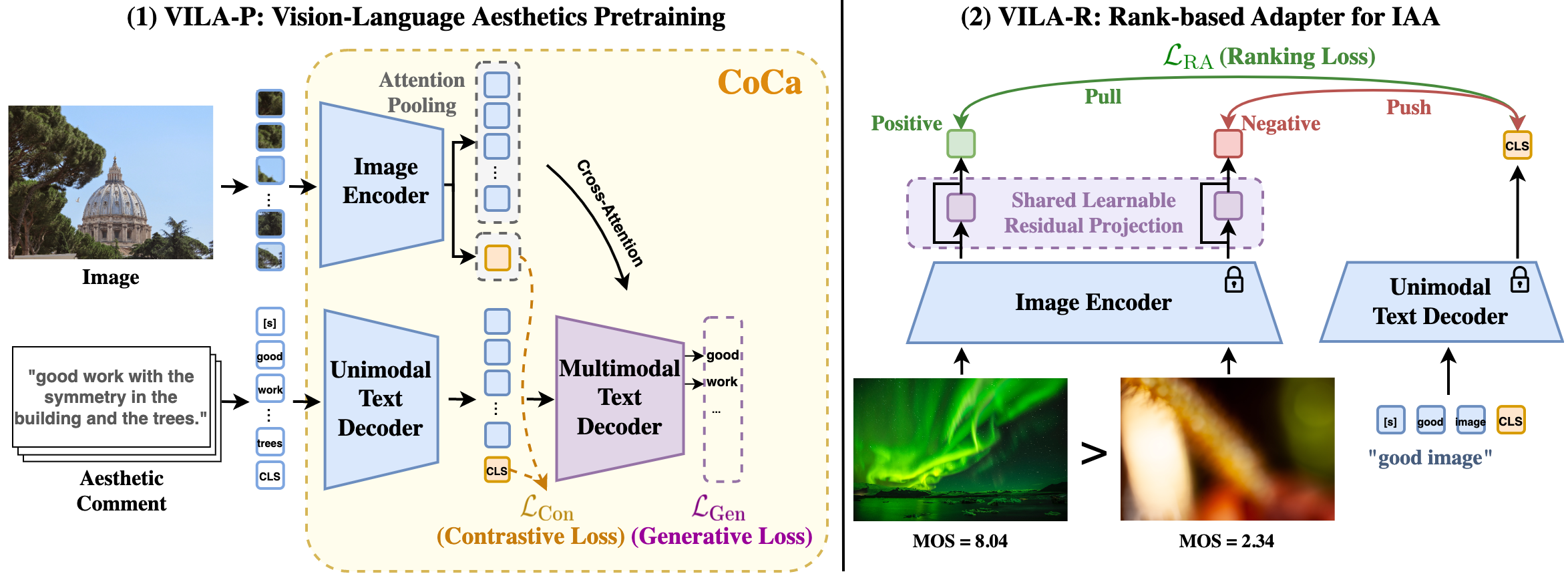}\vspace{-0.5mm}
    \caption{Our proposed vision-language aesthetic (\ours) framework contains two parts: (1) \ours-P: pretraining a vision-language model using images and user comments on aesthetics, and (2) \ours-R:  a rank-based adapter that efficiently adapts the frozen pretrained model to score-based IAA with a small amount of tunable parameters (purple block). }\vspace{-2.5mm}
    \label{fig:overview}
\end{figure*}

Using image-comment pairs for aesthetics learning remains largely unexplored. While previous works have leveraged user comments to improve IAA, their approaches differ significantly from ours.
For example, ~\cite{zhou2016joint,hii2017multigap,zhang2021mscan} proposed to aggregate visual and comment features, yet they require both the image and comment as inputs during inference. This requirement makes it difficult to use such methods in real-world settings where images may not always be accompanied by comments. To mitigate this, Niu \etal \cite{niu2022comment} proposed to use the LDA topics~\cite{blei2003latent} from the comments as pseudo labels to guide image representation learning. However, the simplification of comments into topics may result in a loss of valuable contextual information. Therefore, we are motivated to explore other strategies for utilizing raw comments to extract richer aesthetic textual information. 

In this paper, we present a novel two-stage \textbf{VI}sion-\textbf{L}anguage \textbf{A}esthetics (\textbf{VILA}) learning framework incorporating image-text pretraining. Our goal is to develop a model that can effectively generalize to multiple downstream aesthetic tasks (Fig.~\ref{fig:teaser}). In the first \textbf{P}retraining stage, we learn an image-text model (\textbf{\ours-P}) by employing contrastive and text sequence generation objectives, enbaling us to fully leverage fine-grained knowledge from aesthetic image-comment pairs.
Our approach is motivated by recent advancements in vision-language models, such as CLIP~\cite{radford2021learning}, ALIGN~\cite{jia2021scaling}, and CoCa~\cite{yu2022coca}, which exhibit impressive performance and generalization ability across multiple tasks. These models align vision and language feature spaces to capture the rich semantic information.
However, these models are typically pretrained on general image-text pairs from the web, which can result in under-representation of aesthetic-related information. Our experimental results indicate that such generally pretrained vision-language models underperform on aesthetic tasks (Sec.~\ref{sec:results:mos}). As a solution, we propose the adoption of vision-language pretraining on aesthetic image-comment pairs from photograph sharing websites. To the best of our knowledge, our work is the first to explore the use of image-comment pairs in vision-language pretraining for aesthetics learning.

After pretraining \ours-P on image-comment pairs, we finetune it for downstream score-based IAA tasks using a lightweight \textbf{R}ank-based adapter (\textbf{\ours-R}). This adapter involves adding feature residuals to the frozen image embeddings to move images with high aesthetic quality closer to the anchor text ``good image," and images with low aesthetic quality away from it. This method can effectively rank images based on human rated preferences. With 0.1\% tunable parameters, our model outperforms previous works on IAA correlation metrics over the AVA dataset~\cite{murray2012ava}.

Our proposed \ours{} is capable of tackling multiple aesthetic-related tasks beyond score-based IAA (Fig.~\ref{fig:teaser}). Not only can it generate high-quality aesthetic comments, but it also exhibits impressive zero-shot learning (ZSL) capabilities for aesthetic style classification and quality analysis. Using text queries such as ``good image" and ``bad image" to compare images, our ZSL model outperforms supervised learning models like NIMA~\cite{talebi2018nima} which requires labor-intensive ratings as ground truth. This highlights the potential of learning rich image aesthetic concepts without relying on human-labeled data, thereby significantly reducing data collection costs.

We summarize the contributions of our work as follows:

\begin{itemize}[nolistsep,noitemsep]
\setlength
    \item We propose a vision-language aesthetic learning framework (\ours) for learning rich image aesthetic features using image-comment pairs. 
    \item We design a novel rank-based module to adapt the model to downstream IAA tasks without perturbing the pretrained weights, effectively learning the aesthetic quality concepts with minimal additional parameters.
    \item Our pretrained aesthetic model outperforms prior works for aesthetic captioning on the AVA-Captions~\cite{ghosal2019aesthetic} dataset. Even without any supervised labels, our zero-shot model achieves 69\% mAP on the AVA-Style~\cite{murray2012ava} dataset and 0.657 SRCC on the AVA dataset~\cite{murray2012ava}, outperforming many supervised approaches. With the proposed adapter and a small number of tunable parameters, our method further achieves state-of-the-art performance on AVA.
\end{itemize}

%% file: 2-related-work.tex
\section{Related Work}
\label{sec:related}

\noindent\textbf{Image Aesthetic Assessment} has a wide range of applications such as search, ranking, and recommendation. Unlike the technical quality assessment~\cite{koniq10k,Ying_2020_CVPR,Fang_2020_CVPR} which focuses on image distortion, cropping, or noise, IAA aims to measure the aesthetic quality.
During the deep learning era, works such as \cite{murray2012ava,karayev2013recognizing,tang2013content,Ren_2017_ICCV,lee2018photographic,Yang_2022_CVPR,he_2022_ijcai} focused on data-driven methods and collected large-scale datasets containing images and human ratings.
Based on these datasets, \cite{kong2016photo} built a ranking-based model, while \cite{murray2017deep,talebi2018nima,zeng2019unified} proposed to approximate the groundtruth score distributions. 
Different from these works, our model benefits from the image-text pretraining framework that has rarely been explored in IAA.

Additional supervision in IAA has been explored in works such as \cite{zhou2016joint,wang2019neural}, where natural language annotations were introduced in their curated datasets. However, these methods either treat IAA as one of multiple parallel tasks~\cite{wang2019neural,niu2022comment}, do not generate quality related outputs~\cite{zhou2016joint,wang2019neural}, or require both image and comment at inference time~\cite{zhou2016joint, hii2017multigap, zhang2021mscan}. In contrast, our model leverages user comments to learn meaningful aesthetic representations using contrastive and generative targets, and the learned image model can be used independently without text input.

Moreover, various studies have focused on network design to preserve high-resolution aesthetic information for IAA, such as CNN-based methods \cite{Mai_2016_CVPR,Hosu_2019_CVPR,Chen_2020_CVPR} that reduce the negative effects of cropping and resizing, and transformer architectures \cite{Ke_2021_ICCV,ghosal2022image} that treat input image as visual tokens and support variable-length sequences, preserving image resolution and aspect ratios. Our method achieves state-of-the-art results with a fixed $224\times224$ input without considering original resolution and aspect ratios, and we believe that these related methods could further enhance our model and be incorporated in future work.

\noindent\textbf{Image-Text Pretraining} utilizes the fact that paired image and text are correlated. Initially, contrastive learning was used to draw image representation and aligned text representation closer \cite{frome2013devise,kiros2014unifying,faghri2018vse++}.
Later, self-supervised learning objectives were explored, such as masked region reconstruction, masked object prediction, word region alignment \cite{chen2019uniter,lu2019vilbert,li2019visualbert,su2019vlbert,tan-bansal-2019-lxmert}. These early models used off-the-shelf visual detectors, which limited their generalization to large-scale pretraining. The introduction of ViT~\cite{dosovitskiy2020image} enabled end-to-end multimodal transformer-based methods \cite{kim2021vilt,wang2021vlmo} for large-scale vision-language pretraining. Recently, several methods such as  CLIP~\cite{radford2021learning}, ALIGN~\cite{jia2021scaling}, and CoCa~\cite{yu2022coca} have proposed image-text foundation models trained on large-scale image-text corpus~\cite{jia2021scaling,Zhai_2022_CVPR}. These methods adopted general pretraining using billions of image-text pairs from the web, and showed impressive results on various tasks such as retrieval, classification, and captioning. Concurrent works \cite{hentschel2022clip, wang2022exploring} have shown the benefit of using such generally pretrained CLIP features for aesthetics learning. However, due to the sparsity of aesthetics-related image-text pairs on the web, aesthetic information gets diluted in such general pretraining process. To address this, we propose the aesthetics pretraining on image-comment pairs to further enhance aesthetics information. Our model is based on the CoCa~\cite{yu2022coca} architecture, with a novel rank-based adapter module designed for IAA to learn relative aesthetic quality with minimal tunable parameters. The rank-based adapter optimizes only a small set of learnable parameters, avoiding catastrophic forgetting~\cite{french1999catastrophic,kirkpatrick2017overcoming} while retaining the rich knowledge from the pretrained model.

%% file: 3-method.tex
\section{Image Aesthetics Pretraining using CoCa}
\label{sec:method:image_text}
\noindent In this section, we present our approach to pretrain the image aesthetic model \ours-P. Our goal in the pretraining stage is to learn powerful multimodal representations for image aesthetics in a self-supervised manner, using both images and their associated user comments. 

Without loss of generality, we adopt the CoCa \cite{yu2022coca} architecture, which combines contrastive learning and image-to-caption generation in a single framework.  Our approach is generally applicable to broader vision-language pretraining models. Fig.~\ref{fig:overview} (1) provides an overview of our pretraining architecture for \ours-P.

\subsection{Preliminary of CoCa}
\noindent CoCa contains an image encoder, a unimodal text decoder, and a multimodal text decoder. The image encoder produces an image representation, while the unimodal text decoder generates a text representation with an appended \texttt{[CLS]} token. These two representations are aligned using a contrastive objective.  The multimodal text decoder generates captions by cross-attending to the image features.

\noindent \textbf{Encoding Image:} The image encoder is in the form of a Vision Transformer~\cite{dosovitskiy2020image}, which splits an image into patches and treats them as tokens. The patches are then projected to $D$-dimensional features and fed to the transformer blocks to generate a sequence of visual embeddings $\vect{V} = \{\vect{v}_1, ..., \vect{v}_K\}$, where $K$ is the number of visual tokens.

\noindent \textbf{Encoding Text: } The text is first tokenized into a sequence of tokens, with each token  mapped to a $D$-dimensional word embedding vector. A \texttt{[CLS]} token is appended to the sequence, and the sequence is passed through transformer layers to generate the unimodal text representation $\vect{W} = \{\vect{w}_1, ..., \vect{w}_L, \vect{w}_{cls} \}$, where $\vect{w}_{cls}$ is output of the \texttt{[CLS]} token, and $L$ is the number of text tokens. The transformer text decoder layers are trained with causally-masked self-attention for the captioning objective, which prevents tokens from attending to future tokens. The learnable token $\vect{w}_{cls}$ is used as the contrastive text embedding.

\noindent \textbf{Contrastive Learning Objective: } The two unimodal encoding modules are jointly optimized by a contrastive target which tries to align the image-text pairs:
\vspace{-1mm}
\begin{align}
    \label{eq:contrastive_loss}
    \begin{split}
        \mathcal{L}^{i2t}_{\text{Con}} &= -\frac{1}{N} \Big(\sum_i^N \log \frac{\exp (\vect{x}_i^\top\vect{y}_i)/\tau}{\sum_{j=1}^N\exp(\vect{x}_i^\top\vect{y}_j/\tau)}\Big) \\
        \mathcal{L}^{t2i}_{\text{Con}} &= -\frac{1}{N} \Big(\sum_i^N \log \frac{\exp (\vect{y}_i^\top\vect{x}_i)/\tau}{\sum_{j=1}^N\exp(\vect{y}_i^\top\vect{x}_j/\tau)} \Big) \\
        \mathcal{L}_{\text{Con}} &= \mathcal{L}^{i2t}_{\text{Con}} + \mathcal{L}^{t2i}_{\text{Con}}
    \end{split}
\end{align}
\vspace{-1mm}

\noindent $\vect{x}_i$ and $\vect{y}_i$ are the normalized contrastive embeddings of the $i$-th image and text in the batch. $L^{i2t}_{\text{Con}}$ is the image-to-text contrastive loss and $ L^{t2i}_{\text{Con}}$ is the text-to-image counterpart, $\tau$ is the learnable temperature, $N$ is the batch size.

\noindent \textbf{Generative Learning Objective: } For captioning, the multimodal text decoder learns to maximize the likelihood of generating the paired text conditioned on visual features in an autoregressive manner:
\vspace{-1mm}
$$
\mathcal{L}_{\text{Gen}} = -\sum_{t=1} ^ L \log P (\vect{w}_t | \vect{w}_{< t}, \vect{V}).
$$
\vspace{-2mm}

\noindent\textbf{Cotraining Contrastive and Generative Objective: } \noindent To cotrain the two targets, two task-specific attentional pooling layers \cite{lee2019set} are added on top of the image encoder to generate a contrastive image representation and a generative image representation. The pretraining objective is a weighted sum of the contrastive loss and the generative loss, using hyper-parameters $\alpha$ and $\beta$:
\vspace{-1mm}
\begin{align}
\mathcal{L} = \alpha \mathcal{L}_{\text{Con}} + \beta \mathcal{L}_{\text{Gen}}.
\end{align}
\vspace{-4mm}

\subsection{Vision-Language Pretraining for Aesthetics}
\noindent Vision-language pretraining methods require large-scale data to learn the complex dynamics between visual and textual information. Many of these methods are trained on large proprietary datasets~\cite{radford2021learning, jia2021scaling} with image-text pairs crawled from the web. While this  general pretraining strategy has proven useful for tasks such as image classification and retrieval, it is limited in its ability to represent aesthetic-related information due to the under-representation of such information on the web. Consequently, the aesthetic information gets diluted in the vast amount of pretraining data. To address this limitation, we propose a two-stage pretraining approach that involves initializing the model with a generally pretrained image-text model and then further pretraining it on aesthetic image-comment pairs. For general pretraining, we use a 650M filtered subset of the openly available LAION-5B-English~\cite{schuhmann2022laion} dataset. For aesthetic pretraining, we use the AVA-Captions dataset~\cite{ghosal2019aesthetic} which is currently the largest available dataset for aesthetic comments. Each image in AVA-Captions is associated with one or more user comments that provide informative insights into different aesthetic aspects of the image. We randomly sample one comment for each image to construct image-comment pairs during training.

In contrast to traditional supervised learning with pre-defined labels or categories, vision-language pretraining enables learning of open-set aesthetic concepts through noisy image-comment pairs. This results in visual and textual representations that encompass a wider range of aesthetic concepts, enhancing transferability to downstream tasks.

\section{Adapting Vision-Language Model for IAA}
\label{sec:method:regression_prompts}
\noindent The pretrained model \ours{}-P contains extensive multimodal aesthetic information, enabling it to perform zero-shot aesthetic tasks and to even outperform supervised models (Sec~\ref{sec:results:mos} and Sec~\ref{sec:results:ava_comments}). In this section, we aim to further enhance the model's performance for IAA tasks using the mean-opinion-score (MOS) labels. Finetuning the entire model is computationally expensive and can harm the pretrained model's zero-shot and captioning capability. Therefore, we propose a lightweight rank-based adapter module that adapts the pretrained vision-language model to downstream IAA tasks while keeping the image and text backbone frozen with only a few tunable parameters. The adapter module allows the model to retain the benefits of the pretrained backbone, while leveraging the rich aesthetic textual information for IAA tasks. Fig.~\ref{fig:overview} (2) depicts the overview of the adapter module, and we refer to the resulting model as \ours-R.

\subsection{Image Aesthetic Assessment Formulation}
\noindent The goal of IAA is to predict the aesthetic score for a given image. We focus on the case where the image is represented by the frozen image embedding extracted by the image encoder in \ours-P. Formally, 
\vspace{-1mm}
\begin{align}
\vect{v} &= E(\vect{I}, \vect{\theta}_{frozen}), \\
r &= F(\vect{v}, \vect{\gamma}),
\end{align}
\noindent where $\vect{I}$ is the input image,  $\vect{v}$ is the image features extracted using image encoder $E$ with its frozen pretrained weights $\vect{\theta}_{frozen}$.  $F$ is the IAA scoring model with parameters $\vect{\gamma}$, and $r$ is the predicted aesthetic score.

During training, given two images represented by $\vect{v}_i$ and $\vect{v}_j$, and their corresponding MOS labels $l_i$ and $l_j$, the IAA model output $r_i$ and $r_j$ are trained to respect the order of $l_i$ and $l_j$. The performance of the proposed model $F$ is evaluated by the correlation between $r$ and $l$.

To obtain an effective $F$ with few parameters, we draw inspiration from the ZSL setting where no parameter tuning is required. Since the cosine similarity between paired image-text is maximized by the contrastive pretraining objective (Eq.~\ref{eq:contrastive_loss}), we can use the cosine similarity between the contrastive image embedding $\vect{v}$ and the text embedding $\vect{w}$ as a measure of how much the image aligns with the textual concept. By using text as ``prompts", we can effectively score images for the textual concept (\emph{e.g.}, whether they are ``good image''). Our preliminary study shows that using text prompts for IAA scoring results in a correlation of over $0.6$, suggesting  that the text decoder in \ours-P contains useful information about what constitutes a visually pleasing image.  We aim to utilize this information as an anchor to further enhance the model's IAA ranking capability by designing a lightweight rank-based adapter module.

\subsection{Rank-based Adapter Module}
 
\noindent The pretraining process, which includes contrastive and generative objectives, captures rich textual concepts related to aesthetically pleasing images in the text decoder, and embeds them in the same latent space as the image. Therefore, we can make slight adjustments to the image embedding to improve its alignment with these textual concepts. Concretely, we propose using the frozen text embedding of ``good image" as an anchor to score images, and optimize the relative ranking between two images according to their MOS labels by adjusting their image representations. This is illustrated in Fig.~\ref{fig:overview} (2).

Let  $\vect{v}$ represent the unnormalized contrastive image embedding from the frozen \ours-P image encoder. To obtain the rank-adjusted image embedding $\Tilde{\vect{v}}$, we add a learnable residual represented by $\vect{H} \in \mathbb{R}^{D\times D}$ and normalize the output as follows:
\vspace{-1mm}
\begin{align}
\Tilde{\vect{v}} = \mathrm{normalize}(\vect{v}^\top \vect{H} + \vect{v}),
\end{align}

Next, we use ``good image" as the prompt, and extract its normalized frozen text embedding $\vect{w}_p$ from the \texttt{[CLS]} position of the unimodal text decoder. The cosine similarity between the rank-adjusted image embedding $\Tilde{\vect{v}}$ and the anchor $\vect{w}_p$ is used as the predicted IAA score for ranking:
\vspace{-1mm}
\begin{align}
r = \Tilde{\vect{v}}^\top \vect{w}_p
\end{align}

To optimize the relative ranking between two images, we use $\vect{w}_p$ as the anchor and optimize the triplet ranking loss $\mathcal{L}_{\text{RA}}$ for a pair of input images:

\vspace{-1mm}
\begin{align}
\mathcal{L}_{\text{RA}} = \frac{1}{P} \sum_{i, j, i\neq j, l_i > l_j } \max \Big(0, m -  \Tilde{\vect{v}}_i^\top \vect{w}_p + \Tilde{\vect{v}}_j^\top \vect{w}_p\Big)
\end{align}

\noindent $m$  is the margin hyper-parameter with default value $0.1$. The positive sample $\Tilde{\vect{v}}_i$ corresponds to the image with a higher MOS label $l_i$, and the negative sample $\Tilde{\vect{v}}_j$ corresponds to the image with a lower MOS label $l_j$. The ranking loss ensures that the  similarity between the positive sample  and the ``good image" anchor is greater than that of the negative sample, effectively ranking the images according to its aesthetic ratings. The only tunable parameter is $\vect{H}$ with $D^2$ parameters, about 0.1\% of the total parameters in \ours-P. 

It is worth noting that the frozen text embedding $\vect{w}_p$ can be exported for training and inference without the text backbone. Therefore, the final IAA model has the same computational and storage as a single image-encoder-only model, and it only needs the image as input for IAA inference.

%% file: 4-experiment.tex
\section{Experiments}
\label{sec:results}
\subsection{Datasets}
\label{sec:results:datasets}

\noindent \textbf{LAION-5B-English-Filtered} is a 650M subset from the English split in LAION-5B~\cite{schuhmann2022laion}, which is currently the largest publicly available dataset with 5B CLIP-filtered image-text pairs. The filtered subset is obtained by removing non-informative or bad data, such as poorly formatted text, bad image size or aspect ratio, and poor image content. We use this subset for general image-text pretraining.

\noindent \textbf{AVA Dataset}~\cite{murray2012ava} is a widely-used IAA benchmark originating from the DPChallenge website. It consists of over 250,000 images with user voting scores ranging from 1 to 10. We evaluate the IAA performance of our model on the available 19,928 AVA test images, reporting Spearman rank order correlation coefficient (SRCC) and Pearson linear correlation coefficient (PLCC) metrics.

\noindent \textbf{AVA-Captions}~\cite{ghosal2019aesthetic} dataset is a collection of user comments for the AVA images, crawled from the DPChallenge website, with basic text filtering applied. It contains 230k images and 1.5M captions, with an average of 5 comments per image. To avoid potential data leakage, we strictly follow the official data split of both AVA and AVA-Captions, excluding both test sets from training, resulting in a training dataset with 212,585 images paired with 1.2M captions. We evaluate the aesthetic comment generation quality of our model on 9,361 AVA-Captions test images, reporting BLEU~\cite{papineni2002bleu}, ROUGE~\cite{rouge2004package}, and CIDEr~\cite{vedantam2015cider} scores.

\noindent \textbf{AVA-Style}~\cite{murray2012ava} contains images with 14 photographic style labels. We use the 2,809 testing images to assess the zero-shot aesthetic style classification capability of our pretrained model.

\subsection{Implementation Details}
\label{sec:results:details}


\noindent We use CoCa-Base, the smallest variant of CoCa~\cite{yu2022coca}. It contains a ViT-B/16~\cite{dosovitskiy2020image} image encoder with 12 transformer~\cite{vaswani2017attention} layers, hidden dimension $D=768$, and MLP size $3072$. The image resolution is set to $224\times224$ with a patch size of  $16\times 16$, resulting in $K=196$ image tokens. Data augmentation during training includes random horizontal flipping and random cropping from $272\times272$. The unimodal text decoder consists of 6 transformer layers with the same hidden dimension and MLP size, while the multimodal text decoder consists of another 6 transformer layers. The maximum text length is set to $64$ during training.  For LAION pretraining, we train with 4096 batch size for 500k steps, using 5e-4 learning rate with linear decay to zero, and $0.01$ weight decay. For image aesthetic pretraining on AVA-Captions, we train with 128 batch size for 500k steps, using 1e-5 learning rate with linear decay to zero, and $0.04$ weight decay. We set contrastive loss weight $\alpha=1$ and generative loss weight $\beta=2$. A trainable temperature $\tau$ with an initial value of $0.07$ is used for the contrastive loss, following \cite{yu2022coca, jia2021scaling}. To finetune the rank-based adapter on AVA, we train with 128 batch size for 30k steps using 1e-5 learning rate with linear decay to zero, and $0.01$ weight decay. All experiments use the Adafactor~\cite{shazeer2018adafactor} optimizer with $\beta_1=0.9, \beta_2=0.999$, and are conducted on TPUv3.

\subsection{AVA Image Aesthetic Assessment}
\label{sec:results:mos}

\input{Tables/1-mos-prediction}

\noindent \textbf{Comparing to SOTA.} Tab.~\ref{tab:mos_prediction} shows our results on the AVA dataset. The first group shows the baselines including the ranking method \cite{kong2016photo}, distribution matching based approaches \cite{murray2017deep,talebi2018nima,zeng2019unified}, customized neural networks \cite{Hosu_2019_CVPR,Chen_2020_CVPR,Ke_2021_ICCV,ghosal2022image,tu2022maxvit}, and semantic-aware methods \cite{he_2022_ijcai,niu2022comment,hentschel2022clip}.
Our approach \ours{}-R achieves the best performance overall and outperforms the current SOTA GAT$_{\times3}$-GATP~\cite{ghosal2022image} by 1.6\% and 1.3\% in terms of SRCC (0.774 vs 0.762) and PLCC (0.774 vs 0.764), respectively. Moreover, our method uses a lower resolution of 224$\times$224 while other methods may benefit from the larger inputs. For example, MUSIQ~\cite{Ke_2021_ICCV} uses the full-size image and two additional resolutions, yet it  underperforms our model.  Hentschel \etal \cite{hentschel2022clip} utilize frozen CLIP features for learning image aesthetics, and \ours{}-R outperforms their approach, which shows the additional benefit of the proposed aesthetic pretraining.

\noindent \textbf{Zero-shot Learning (ZSL) for IAA.} The second group in Tab.~\ref{tab:mos_prediction} shows the results of using our image-text pretrained model \ours-P (Sec.~\ref{sec:method:image_text}) for zero-shot IAA. We utilize the cosine similarity between the contrastive image and text embeddings for these experiments. In the single prompt setting, we compute the cosine  similarity between the image and a single pair of prompts (``good image", ``bad image"), and use the softmax normalized output for ``good image" as the ZSL score for IAA. For ensemble prompts, we use an average ensemble of six pairs of prompts, each consisting of ``good" or ``bad" plus ``image", ``lighting", ``composition", ``foreground", ``background", and ``content" (see supplementary material). Notably, without any human label supervision, our ZSL model (SRCC 0.657, PLCC 0.663) has already outperformed several supervised baselines such as Kong \etal~\cite{kong2016photo}, NIMA~\cite{talebi2018nima}, and AFDC + SPP~\cite{Chen_2020_CVPR}. These observations demonstrate the potential of leveraging unlabelled user comments for IAA, significantly reducing human labeling costs.

\input{Tables/2-ablation-pretraining}

\input{Tables/3-ablation-adapter}

\noindent \textbf{Effects of image-text pretraining.} Tab.~\ref{tab:ava_captions_pretraining} presents an ablation study to validate the effectiveness of the proposed image-text pretraining. We conduct the general pretraining and aesthetic pretraining on the LAION~\cite{schuhmann2022laion} subset and AVA-Captions~\cite{ghosal2019aesthetic}, respectively. With only the general pretraining, the model has suboptimal performance on the IAA task, verifying the assumption that image aesthetic information gets diluted by the vast amount of unrelated data from the web. Adding aesthetic pretraining greatly improves model performance in both zero-shot and finetuned settings. Both general and aesthetic pretraining have a significant positive impact on the final IAA task predictions. Regardless of the pretraining schema, the proposed rank-based adapter enhances the model's IAA performance with minimally tuned parameters.

\noindent \textbf{Effectiveness of the proposed rank-based adapter.} Tab.~\ref{tab:abaltion_adapter} shows an ablation study for the proposed rank-based adapter (Sec.~\ref{sec:method:regression_prompts}). We compare  different options for adapting the frozen \ours-P to downstream score-based IAA. The first group shows regression baselines that predict either the single MOS score using a L2 loss or the distribution of MOS scores using EMD loss~\cite{talebi2018nima}. \ours-R outperforms both of them, showing the effectiveness of a rank-based target. In the second group, we ablate the components in the proposed adapter. ``w/o Text Anchor" denotes using a learnable projection to replace the frozen text prompt embedding $\vect{w}_p$. \ours-R performs better, showing the benefit of using the rich text embedding as a ranking anchor. For  ``w/o Residual", we use a simple learnable projection without the residual,~\ie, $\Tilde{\vect{v}} = \mathrm{normalize}(\vect{v}^\top \vect{H)}$. Its sub-par performance confirms the intuition that we only need to slightly adjust the image embedding, thus learning the residual is easier. The final line shows that \ours-R can be further improved with finetuning the image encoder. However, its gain in performance comes at the cost of disturbing the generic pretrained weights, \eg its ZSL performance on AVA-Style drops from 69.0\% to 26.3\% mAP. \ours-R enables effective IAA adaptation while inheriting the pretrained weights.  

\subsection{AVA-Captions Image-Text Pretraining}
\label{sec:results:ava_comments}

\noindent In this section we aim to verify \ours-P model learns meaningful representations that are   generalizable to other tasks. We evaluate its performance on zero-shot style classification and the quality of its generated aesthetic comments.

\noindent \textbf{Zero-shot Style Classification.}
To demonstrate that \ours{}-P captures diverse aesthetic aspects such as composition, color, and style, we evaluate its ZSL performance on the AVA-Style test set. We manually curate text prompts based on the 14 class names, and use the cosine similarities to approximate the probability that an image involves specific styles (see supplementary material). Tab.~\ref{tab:style_prediction} shows the results. The first group contains supervised methods trained on 11k images with style annotations. Without such supervision, \ours-P achieves 69.0\% ZSL mAP,  outperforming many supervised methods such as MNet~\cite{sun2017convolution} (65.5\%) and Lu \etal \cite{lu2015deep} (64.1\%). This demonstrates the ability of the proposed framework to learn open-set aesthetic information without human labelling. Tab.~\ref{tab:style_prediction} also shows that the performance of the model trained only with general pretraining is much lower than that with aesthetic pretraining.  This again verifies that the proposed aesthetic pretraining is necessary for capturing rich aesthetic information.

\input{Tables/4-style-prediction}

\noindent \textbf{AVA Comments Generation.} We evaluate the captioning performance of \ours-P on AVA-Captions test set, and the results are shown in Tab.~\ref{tab:comment_generation}. Our method outperforms CWS~\cite{ghosal2019aesthetic} and Yeo \etal~\cite{yeo2021generating} in terms of BLEU-2, BLEU-3, BLEU-4, ROUGE and CIDEr. Although our method has a slightly lower BLEU-1 than CWS, it is important to note that BLEU-1 only measures precision of unigram, while and higher order BLEU scores (BLEU-2, BLEU-3, BLEU-4) place more emphasis on the fluency of generated sentences. Moreover, our method's superior ROUGE and CIDEr scores indicates that our model generates more semantically  similar sentences to the real user comments.

\input{Tables/5-captioning}

\noindent \textbf{Qualitative Examples.}
To properly credit our image sources, we choose to display images from the KonIQ-10k~\cite{koniq10k} dataset instead of the AVA dataset for illustration in this section. The image sources are provided in supplementary material. Fig.~\ref{fig:bad_vs_good} depicts the top-5 images retrieved by text queries ``Bad photo'' and ``Good photo'' on KonIQ-10k. For ``Bad photo'', the retrieved results exhibit poor  lighting, bad composition and meaningless content. In contrast, the ``Good photo'' group has noticeably better aesthetic quality. These examples provide qualitative evidence of the aesthetic knowledge captured by the pretrained model.

\begin{figure}[t]
    \centering
    \vspace{-2mm}
    \includegraphics[width=1\linewidth]{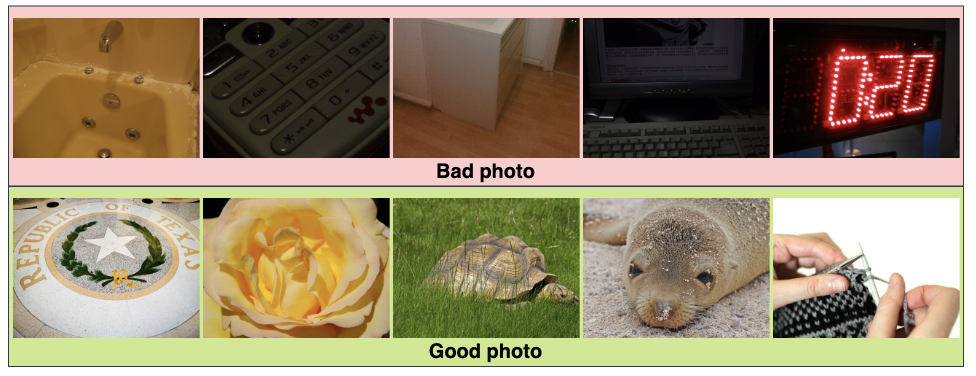}
    \caption{Top 5 images retrieved with ``bad photo'', ``good photo'' on KonIQ-10k~\cite{koniq10k}. See supplementary material for image sources.}
    \label{fig:bad_vs_good}
    \vspace{-4mm}
\end{figure}

\begin{figure}[t]
    \centering
    \includegraphics[width=1\linewidth]{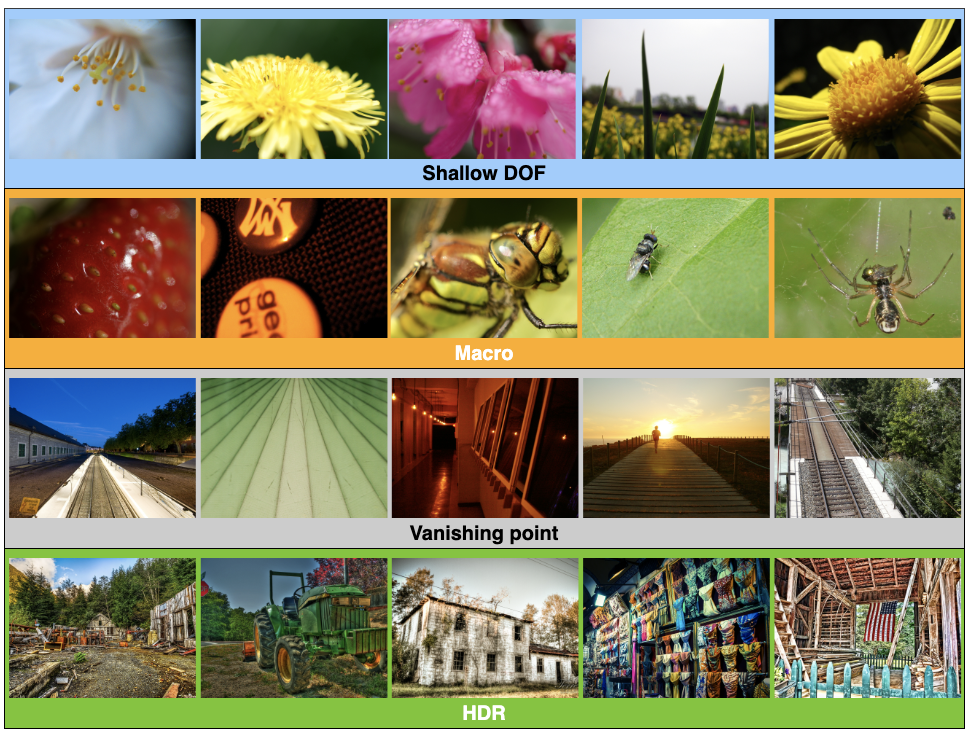}
    \caption{Top 5 images retrieved using AVA-Style class names on KonIQ-10k~\cite{koniq10k}. To give proper attribution to image sources, we choose to showcase images from the KonIQ-10k dataset instead of the AVA dataset. See supplementary material for image sources.}
    \label{fig:style_qualitative_examples} 
    \vspace{-3mm}
\end{figure}

Fig.~\ref{fig:style_qualitative_examples} illustrates the AVA-Style predictions of \ours{} by visualizing the top-5 images retrieved using style class names on KonIQ-10k. This provides a qualitative demonstration of the aesthetic information captured by \ours{}. Results show that the aesthetic pretraining on image-comment pairs has helped the model to understand low-level aesthetic attributes quite well. For example, the learned model understands that ``Macro'' is a visual concept that captures finer details, regardless of the semantic objects, such as strawberry or insects. Another example is ``HDR'', for which all retrieved photos have high dynamic range while portraying  different semantic objects such as buildings and cars.

\begin{figure}[t]
    \centering
    \includegraphics[width=1\linewidth]{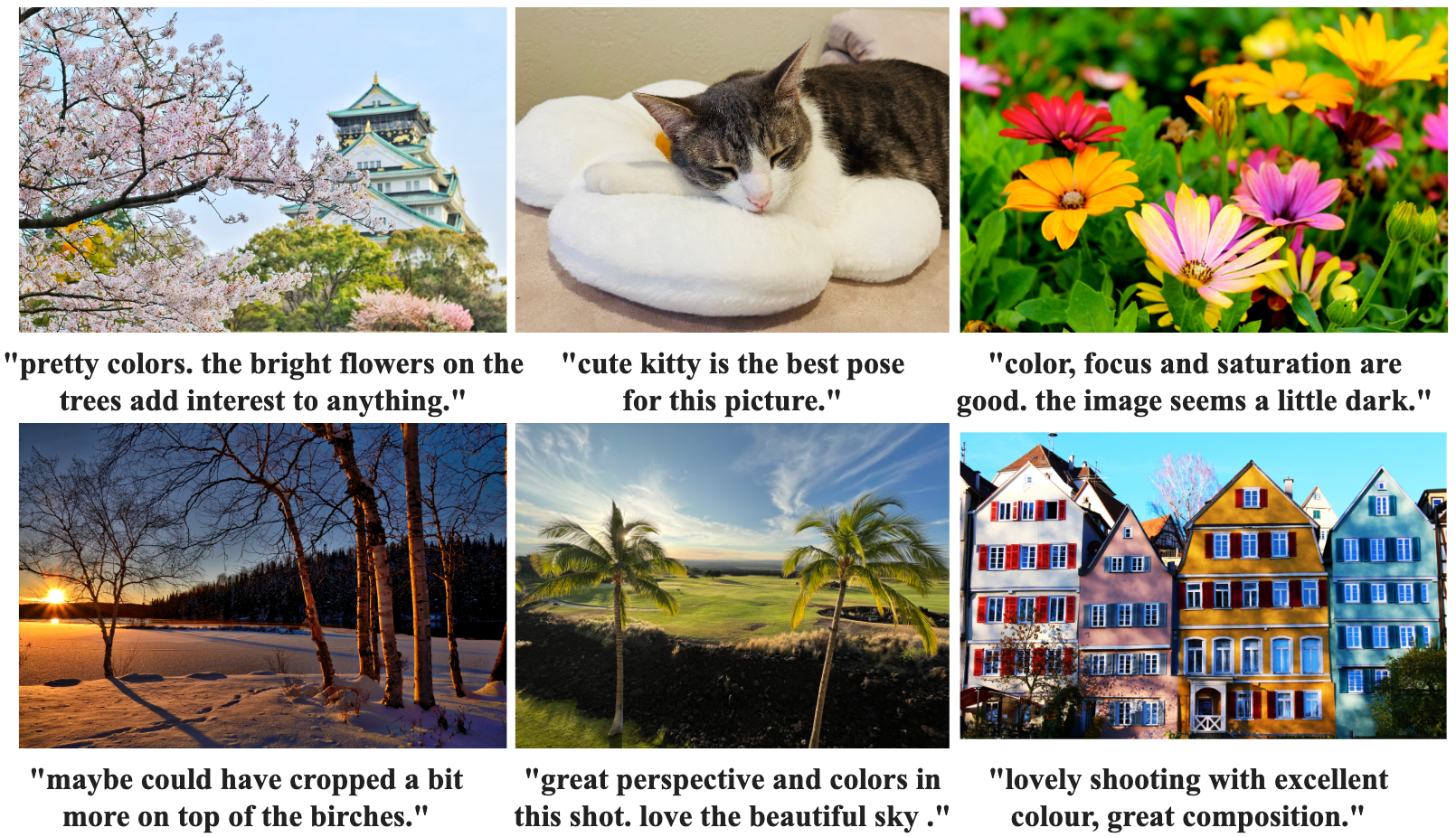}
    \vspace{-3mm}
    \caption{Aesthetic comments generated by \ours.}
    \label{fig:caption_qualitative_examples} 
    \vspace{-4mm}
\end{figure}

Fig.~\ref{fig:caption_qualitative_examples} shows aesthetic comments generated by \ours{}. The model is capable of generating diverse captions conditioned on the images, mentioning attributes such as ``color'', ``saturation'' and ``persepective''. In addition, it even includes critiques about the cropping of the image, which aligns with our aesthetic perspective.

%% file: Tables/1-mos-prediction.tex
\begin{table}[t]
    \centering
    \small
    \begin{tabular}{lcccc}
    \toprule
        Method &SRCC &PLCC \\
    \midrule
        Kong \etal \cite{kong2016photo}                 &0.558 &- \\
        NIMA (Inception-v2) \cite{talebi2018nima}       &0.612 &0.636 \\
        AFDC + SPP \cite{Chen_2020_CVPR}                &0.649 &0.671 \\
        MaxViT \cite{tu2022maxvit}                      &0.708 &0.745 \\
        AMP \cite{murray2017deep}                       &0.709 &- \\
        Zeng \etal (resnet101) \cite{zeng2019unified}   &0.719 &0.720 \\
        MUSIQ \cite{Ke_2021_ICCV}                       &0.726 &0.738 \\
        Hentschel \etal \cite{hentschel2022clip}        &0.731 &0.741 \\      
        Niu \etal \cite{niu2022comment}                 &0.734 &0.740 \\
        MLSP (Pool-3FC) \cite{Hosu_2019_CVPR}           &0.756 &0.757 \\
        TANet \cite{he_2022_ijcai}                      &0.758 &\second{0.765} \\
        GAT$_{\times3}$-GATP \cite{ghosal2022image}     &\second{0.762} &0.764 \\
    \midrule
        \textbf{Zero-shot Learning} \\
        \ours-P (single prompt)                                   &0.605 &0.617 \\
        \ours-P (ensemble prompts)                                &0.657 &0.663 \\
    \midrule
        \ours-R                                           &\best{0.774} &\best{0.774} \\
    \bottomrule

    \end{tabular}
    \vspace{-1mm}
    \caption{Results on AVA dataset. \best{Blue} and \second{black} numbers in bold represent the best and second best respectively. First group shows baselines, second group shows ZSL results using our model from Sec.~\ref{sec:method:image_text}, final line shows our result combining Sec.~\ref{sec:method:image_text} and Sec.~\ref{sec:method:regression_prompts}.}
    \label{tab:mos_prediction}
    \vspace{-3mm}
\end{table}

%% file: Tables/2-ablation-pretraining.tex
\begin{table}[t]
    \centering
    \setlength\tabcolsep{2pt}
    \small
    \begin{tabularx}{1\linewidth}{c|*{3}{>{\centering\arraybackslash}X}|*{3}{>{\centering\arraybackslash}X}}
    \Xhline{2\arrayrulewidth}
        &\multicolumn{3}{c|}{ZSL Ens. Prompts}
        &\multicolumn{3}{c}{w/ Our Adapter} \\
    \Xhline{1\arrayrulewidth}
        General Pretraining   &\cmark &       &\cmark &\cmark &       &\cmark \\
        Aesthetic Pretraining &       &\cmark &\cmark &       &\cmark &\cmark \\
    \Xhline{1\arrayrulewidth}
        SRCC                  & 0.228 & 0.265 & \textbf{0.657} & 0.746 & 0.566 & \textbf{0.774} \\
        PLCC                  & 0.228 & 0.276 & \textbf{0.663} & 0.750 & 0.575 & \textbf{0.774} \\
    \Xhline{2\arrayrulewidth}
    \end{tabularx}
    \vspace{-1mm}
    \caption{Effects of image-text pretraining on AVA. Different pretraining schema are employed for each column and two settings are reported: 1) ZSL using an ensemble of prompts; 2) further finetuned using our proposed rank-based adapter. }
    \label{tab:ava_captions_pretraining}
    \vspace{-2mm}
\end{table}

%% file: Tables/3-ablation-adapter.tex
\begin{table}[t]
    \centering
    \small
    \begin{tabular}{lcccc}
    \toprule
        Method &SRCC &PLCC \\
    \midrule
        \ours-P w/ L2 Loss                            &0.757 &0.756 \\
        \ours-P w/ EMD Loss~\cite{talebi2018nima}     &0.759 &0.759 \\\midrule
        \ours-R w/o Text Anchor                       &0.763 &0.764 \\
        \ours-R w/o Residual                          &0.766 &0.766 \\
        \ours-R (Ours)                                &\textbf{0.774} &\textbf{0.774} \\\midrule
        \textcolor{lightgray}{VILA-R Finetune Image Encoder }
                                                      & \textcolor{lightgray}{0.780}
                                                      & \textcolor{lightgray}{0.780} \\
    \bottomrule
    \end{tabular}
    \vspace{-1mm}
    \caption{Ablation for the proposed rank-based adapter (Sec.~\ref{sec:method:regression_prompts}) on AVA. First two groups use frozen  pretrained image encoder.}
    \label{tab:abaltion_adapter}
    \vspace{-4mm}
\end{table}

%% file: Tables/4-style-prediction.tex
\begin{table}[t]
    \centering
    \small
    \begin{tabular}{lcc}
    \toprule
    Method &mAP (\%) \\
    \midrule
        Murray \etal \cite{murray2012ava}               &\textcolor{lightgray}{53.9} \\
        Karayev \etal \cite{karayev2013recognizing}     &\textcolor{lightgray}{58.1} \\
        Lu \etal \cite{lu2015deep}                      &\textcolor{lightgray}{64.1} \\
        MNet \cite{sun2017convolution}                  &\textcolor{lightgray}{65.5} \\
        Sal-RGB \cite{ghosal2019geometry}               &\textcolor{lightgray}{71.8} \\
    \midrule
        \textbf{Zero-shot Learning} \\
        General Pretraining (single prompt) 
                                                        &29.3 \\
        General Pretraining (ensemble prompts)
                                                        &32.6 \\
        \ours-P (single prompt)                                     &62.3 \\
        \ours-P (ensemble prompts)                       &\second{69.0} \\
    \bottomrule
    \end{tabular}
    \vspace{-1mm}
    \caption{Results on AVA-Style dataset. We \textcolor{lightgray}{gray} out supervised baselines as they are not directly comparable to our unsupervised model which is not exposed to the training labels. }
    \label{tab:style_prediction}
    \vspace{-3mm}
\end{table}

%% file: Tables/5-captioning.tex
\begin{table}[t]
    \centering
    \setlength\tabcolsep{2.25pt}
    \footnotesize
    \begin{tabular}{lccccccc}
    \toprule
        Method &BLEU-1 &BLEU-2 &BLEU-3 &BLEU-4 &ROUGE &CIDEr \\
    \midrule
        CWS \cite{ghosal2019aesthetic}      &\textbf{0.535} &0.282 &0.150 &0.074 &0.254 &0.059 \\
        Yeo \etal \cite{yeo2021generating}  &0.464 &0.238 &0.122 &0.063 &\textbf{0.262} &0.051 \\
    \midrule
        \ours &0.503 &\textbf{0.288} &\textbf{0.170} &\textbf{0.113} &\textbf{0.262} &\textbf{0.076} \\
    \bottomrule
    \end{tabular}
    \vspace{-2mm}
    \caption{Results on AVA-Captions dataset.}
    \label{tab:comment_generation}
    \vspace{-2mm}
\end{table}

%% file: 5-conclusion.tex
\section{Conclusion}
\label{sec:conclusion}
\noindent We propose a general framework for learning image aesthetics (\ours). By pretraining vision-language models on image-comment pairs from image sharing websites, we enable the model to learn rich aesthetic semantics in a self-supervised manner without the need for expensive labeled data. The resulting pretrained model, \ours-P, exhibits state-of-the-art performance on the AVA-Captions dataset and enables various interesting tasks, including zero-shot learning for IAA, style classification, and retrieval. Our experiments demonstrate that \ours{}-P surpasses many supervised baselines on these tasks with ZSL. To efficiently adapt the pretrained model for  IAA without impairing its powerful zero-shot abilities or damaging the rich representation, we introduce a lightweight rank-based adapter module. By employing the text embedding as an anchor and explicitly modeling the ranking concept, we achieve state-of-the-art IAA performance on the AVA dataset with only a small amount of injected parameters. Although we design the rank-based adapter module for IAA, our method is generally applicable for adapting large-scale visual-language models to other ranking based tasks.

%% file: 6-supplementary.tex
\begin{center}
    \Large \textbf{Supplementary Material}
\end{center}

\section{Effect of Hyper-parameter $\alpha$, $\beta$}
\noindent Tab.~\ref{tab:abaltion_alpha_beta} presents the ablation study on the use of different hyper-parameter $\alpha, \beta$ ratios in Eq. (2) during the pretraining of \ours. The results demonstrate that the setting of $\alpha : \beta = 1: 2$ yields the best performance, which is consistent with the recommended setting in the CoCa~\cite{yu2022coca} paper.

\input{Tables/supp-1-ablation-alpha-beta}

\section{Effect of Margin Hyper-parameter $m$}
\noindent Tab.~\ref{tab:abaltion_margin} presents an ablation study for using different margin hyper-parameter $m$ in Eq. (7) when finetuning \ours-R. The results show that a margin of $m=0.1$ achieves the best performance, and we adopt this value as the default for all other experiments.

\input{Tables/supp-2-ablation-margin}

\section{Effect of Random Sampling Comments}
\noindent During training, we create image-comment pairs by randomly selecting one comment from the available list of comments for an image, if there are multiple comments associated with the same image. Tab.~\ref{tab:abaltion_sampling_comment} shows the effect of such random sampling during aesthetic pretraining. When a fixed comment is used for each image, the AVA ZSL performance drops from 0.663 PLCC to 0.596. Random sampling is an effective approach since different comments may cover different aesthetic aspects of the same image, allowing the model to fully expose itself to diverse and rich aesthetic information in the noisy dataset. This strategy enables the mining of open-set aesthetic concepts automatically.
\input{Tables/supp-3-sampling-comment}

\section{Per-class Evaluation on AVA-Style}
\noindent We show the per-class evaluation on AVA-Style in Tab.~\ref{tab:ava_style_prediction_per_class}, comparing to the same baselines as in our main paper. 
\input{Tables/supp-6-style-prediction}

\section{Details on ZSL for AVA-Style Classification}

\input{Tables/supp-5-style-prompts}
\noindent\textbf{Single prompt.}
In this approach, we use the 14 photographic style names as the language prompts: \{``complementary colors'', ``duo tones'', ``hdr'', ``image grain'', ``light on white'', ``long exposure'', ``macro'', ``motion blur'', ``negative image'', ``rule of thirds'', ``shallow dof'', ``silhouettes'', ``soft focus'', ``vanishing point''\}. The cosine similarity between the prompt text embedding and the image embedding is used as the  prediction score.

\vspace{+3mm}
\noindent\textbf{Ensemble of prompts.} In this approach, we manually curate five sentences/phrases that are frequently mentioned in the AVA-Caption user comments, for each of the styles. These prompts either use synonyms (\emph{e.g.} ``color'' and ``colors'') of the styles or add more text contexts (\emph{e.g.}, ``i like the lines and fading or vanishing''). Tab.~\ref{tab:style_prompts} shows these prompts.

\section{Details on ZSL for IAA}
\noindent To effectively perform zero-shot learning for IAA, we use a pair of prompts with opposite meanings (``good" v.s. ``bad").

\vspace{+3mm}
\noindent\textbf{Single prompt.} In this approach, we use \{``good image", ``bad image"\} as input prompts. Let $\vect{p}_g$ and $\vect{p}_b$ be the normalized unimodal text embedding for the ``good" and ``bad" prompts respectively, $\vect{v}$ be the normalized image contrastive embedding. We compute the cosine similarity and use the softmax normalized score for ``good image" as the final score $r$ for IAA.
\begin{align*}
r = \frac{e^{\vect{v}^\top\vect{p}_g}}{e^{\vect{v}^\top\vect{p}_g} + e^{\vect{v}^\top\vect{p}_b}}
\end{align*}

\noindent\textbf{Ensemble of prompts.} In this approach, we similarly construct six pairs of ``good" v.s. ``bad" prompts for \{``image", ``lighting", ``content", ``background", ``foreground", ``composition"). The second group in Tab.~\ref{tab:iaa_prompts} shows these pairs of prompts. For each pair, we can obtain a score $r_i, i = 1,..., 6$. Then we use the average ensemble of the scores to get the final score $r$ for IAA.
\input{Tables/supp-4-iaa-prompts}

\section{Results on KonIQ-10k}
\input{Tables/supp-7-koniq10k}

\noindent Table~\ref{tab:koniq-results} presents additional results on the image quality dataset KonIQ-10k~\cite{koniq10k}. We adopt the same data split as~\cite{koniq10k} and and employ a batch size of 32 to finetune the rank-based adapter for 30k steps, with a learning rate of 5e-4 and linear decay to zero, and 0.04 weight decay. Our proposed VILA-R outperforms CLIP-IQA$^+$ \cite{wang2022exploring} which trains a prompt tuning module on top of CLIP features. While CLIP features only use general pretraining, VILA-R benefits from the aesthetic pretraining which learns rich perceptual quality information, highlighting the importance of the proposed aesthtic pretraining. Remarkably, with only 0.1\% tunable parameters, VILA-R's performance is competitive with KonCept512 \cite{koniq10k} and MUSIQ \cite{Ke_2021_ICCV}, which rely on much larger resolutions. It is worth noting that KonIQ-10k~\cite{koniq10k} is not solely focused on aesthetics quality, and it includes images with technical quality problems such as compression and blur. There is limited user comments mentioning such aspects on the AVA-Captions dataset. Despite the gap, our model demonstrates competitive performance on KonIQ-10k, showcasing its robustness in capturing the visual appeal of the image across different datasets.

\section{More Qualitative Examples}

\begin{figure}[!t]
    \centering
    \includegraphics[width=8cm]{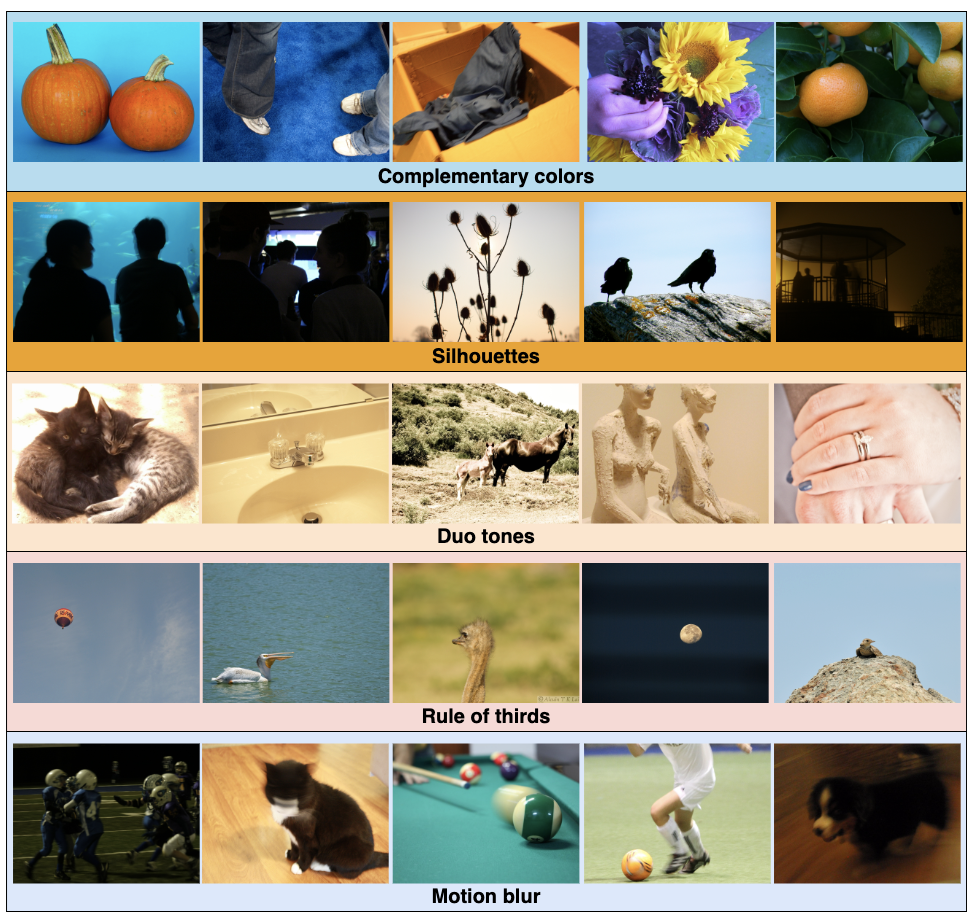}
    \caption{More examples for the top-5 images retrieved using style name query on KonIQ-10k~\cite{koniq10k}. The source of the displayed images are provided in Table~\ref{tab:koniq-attribution}.}
    \label{fig:ava_styles_all}
    \vspace{-2mm}
\end{figure}

\noindent Fig.~\ref{fig:ava_styles_all} displays additional style retrieval results (top-5) on KonIQ-10k~\cite{koniq10k} using AVA-style names as the query. In order to provide clear attribution to the image sources, we have opted to showcase images from the KonIQ-10k dataset instead of the AVA dataset. Attribution to the images are provided in  Table~\ref{tab:koniq-attribution}. Overall, the retrieved results align with our aesthetic perspective. Notably, \ours{} accuratly captures the lighting or color related information. For example, images retrieved for ``Silhouettes" and ``Complementary colors'' accurately depict the corresponding concepts. Additionally, \ours{} recognizes concepts aesthetic concepts like ``Motion blur'' with high accuracy. However, there are also some failure cases where improvements are possible. For example, among the images retrieved using the query ``Rule of thirds", the last three images are centered rather than following the rule of thirds, which may be attributed to the random cropping augmentation during training. Augmentation improvement may help mitigate this issue.  For ``Duo tones", the top retrieved images have a yellowish tone, possibly due to training data bias in the AVA-Captions dataset. Thus, using a more diverse aesthetic pretraining dataset may further enhance the model's performance.

\section{KonIQ-10k Images Attribution}
\noindent In this paper, we display several images from KonIQ-10k~\cite{koniq10k}. The Flickr links and the license information for these images can be found in Table~\ref{tab:koniq-attribution}. We extend our gratitude to the original photographers for sharing their images.

\input{Tables/supp-8-image-attribution}

%% file: Tables/supp-1-ablation-alpha-beta.tex
\begin{table}[!htp]
    \centering
    \footnotesize
    \begin{tabular}{lcccc}
    \toprule
        $\alpha : \beta$ &SRCC &PLCC \\
    \midrule
        $2:1$                            &0.645 &0.651 \\
        $1:1$                            &0.656 &0.661 \\
        $1:2$                             &\textbf{0.657} &\textbf{0.663} \\
    \bottomrule
    \end{tabular}
    \caption{Effect of $\alpha$ and $\beta$ ratios in aesthetic pretraining, comparing zero-shot IAA performance with an ensemble of prompts on the AVA dataset.}
    \label{tab:abaltion_alpha_beta}
\end{table}

%% file: Tables/supp-2-ablation-margin.tex
\begin{table}[!htp]
    \centering
    \footnotesize
    \begin{tabular}{lcccc}
    \toprule
        Margin &SRCC &PLCC \\
    \midrule
        $m=0.01$                            &0.769 &0.769 \\
        $m=0.05$                            &0.770 &0.770 \\
        $m=0.1$                             &\textbf{0.774} &\textbf{0.774} \\   
        $m=0.15$                            &0.772 &0.771 \\
        $m=0.2$                             &0.772 &0.770 \\
    \bottomrule
    \end{tabular}
    \caption{Ablation for different margin hyper-parameter $m$  for the proposed rank-based adapter tuning on AVA dataset.}
    \label{tab:abaltion_margin}
\end{table}

%% file: Tables/supp-3-sampling-comment.tex
\begin{table}[!htp]
    \centering
    \footnotesize
    \begin{tabular}{ccccc}
    \toprule
        Comment Sampling &SRCC &PLCC \\
    \midrule
        Random                             &\textbf{0.657} &\textbf{0.663} \\
        Fixed                             &0.585 &0.596 \\
    \bottomrule
    \end{tabular}
    \caption{Effect of random sampling comment in aesthetic pretraining, comparing zero-shot IAA performance with an ensemble of prompts on the AVA dataset.}
    \label{tab:abaltion_sampling_comment}
\end{table}

%% file: Tables/supp-6-style-prediction.tex
\begin{table*}[ht!]
    \scriptsize
    \centering
    \setlength\tabcolsep{1.2pt} 
    \begin{tabularx}{1.0\linewidth}{l*{14}{>{\centering\arraybackslash}X}|*{1}{>{\centering\arraybackslash}X}}
    \Xhline{2\arrayrulewidth}
        & Compl. Colors
        & Duo tones
        & HDR
        & Image Grain
        & Light On White
        & Long Expos.
        & Macro
        & Motion Blur
        & Negative Image
        & Rule of Thirds
        & Shallow DOF
        & Silhouet.
        & Soft Focus
        & Vanish. Point
        & mAP \\
    \Xhline{1\arrayrulewidth}
        Murray \etal [32]                 & -&-&-&-&-&-&-&-&-&-&-&-&-&-& \textcolor{lightgray}{53.9} \\
        Karayev \etal [18]                & \textcolor{lightgray}{46.9} & \textcolor{lightgray}{67.6} & \textcolor{lightgray}{66.9} & \textcolor{lightgray}{64.7} & \textcolor{lightgray}{90.8} & \textcolor{lightgray}{45.3} & \textcolor{lightgray}{47.8} & \textcolor{lightgray}{47.8} & \textcolor{lightgray}{59.5} & \textcolor{lightgray}{35.2} & \textcolor{lightgray}{62.4} & \textcolor{lightgray}{79.1} & \textcolor{lightgray}{31.2} & \textcolor{lightgray}{68.4} & \textcolor{lightgray}{58.1} \\
        Lu \etal [29]                     & -&-&-&-&-&-&-&-&-&-&-&-&-&-& \textcolor{lightgray}{64.1} \\
        MNet [42]                         & -&-&-&-&-&-&-&-&-&-&-&-&-&-& \textcolor{lightgray}{65.5} \\
        Sal-RGB [10]                      & \textcolor{lightgray}{61.4} & \textcolor{lightgray}{87.6} & \textcolor{lightgray}{72.9} & \textcolor{lightgray}{82.2} & \textcolor{lightgray}{83.0} & \textcolor{lightgray}{61.9} & \textcolor{lightgray}{66.6} & \textcolor{lightgray}{62.0} & \textcolor{lightgray}{87.7} & \textcolor{lightgray}{41.7} & \textcolor{lightgray}{82.4} & \textcolor{lightgray}{93.1} & \textcolor{lightgray}{46.4} & \textcolor{lightgray}{76.8} & \textcolor{lightgray}{71.8} \\
    \Xhline{1\arrayrulewidth}
        \textbf{Zero-shot Learning} \\
        General Pretraining (single prompt)   & 36.5 & 21.0 & 23.9 & 7.1 & 37.0 & 34.6 & 49.9 & 32.8 & 12.7 & 14.3 & 33.5 & 67.9 & 15.6 & 23.4 & 29.3 \\
        General Pretraining (ensemble prompts) & 36.0 & 51.3 & 36.9 & 8.1 & 30.5 & 40.4 & 55.0 & 33.4 & 13.8 & 14.5 & 27.0 & 64.9 & 18.0 & 27.3 & 32.6 \\
        \ours{}-P (single prompt)           & 48.1 & 55.8 & 76.6 & 76.0 & 72.9 & 66.1 & 70.8 & 67.6 & 34.9 & 25.8 & 77.9 & 81.6 & \textbf{51.2} & 67.3 & 62.3 \\
        \ours{}-P (ensemble prompt)         & \textbf{53.6} & \textbf{81.8} & \textbf{79.3} & \textbf{86.7} & \textbf{75.4} & \textbf{69.2} & \textbf{72.9} & \textbf{74.1} & \textbf{58.6} & \textbf{30.9} & \textbf{78.6} & \textbf{85.4} & 51.0 & \textbf{67.8} & \textbf{69.0}\\
    \Xhline{2\arrayrulewidth}
    \end{tabularx}
    \caption{AVA-Style per-class evaluation results. Supervised baselines are shown in \textcolor{lightgray}{gray} color.}
    \label{tab:ava_style_prediction_per_class}
\end{table*}

%% file: Tables/supp-5-style-prompts.tex
\begin{table*}[!htp]
    \centering
    \footnotesize
    \begin{tabular}{cc}
    \toprule
        Style &Prompts \\
    \midrule
        \multirow{5}{*}{Complementary\_Colors}
        & \textit{``complementary colors''} \\
        & \textit{``complementary color''} \\
        & \textit{``great complementary colors''} \\
        & \textit{``great use of complementary colors''} \\
        & \textit{``good use of complementary colors''} \\
    \midrule
        \multirow{5}{*}{Duotones}
        & \textit{``duo tones''} \\
        & \textit{``duotone''} \\
        & \textit{``nice duotone''} \\
        & \textit{``duotone works very well''} \\
        & \textit{``use of duotone''} \\
    \midrule
        \multirow{5}{*}{HDR}
        & \textit{``hdr''} \\
        & \textit{``i like the hdr''} \\
        & \textit{``great job with the hdr''} \\
        & \textit{``hdr done well''} \\
        & \textit{``love the hdr shot''} \\
    \midrule
        \multirow{5}{*}{Image\_Grain}
        & \textit{``image grain''} \\
        & \textit{``i like the image grain''} \\
        & \textit{``nice use of image grain''} \\
        & \textit{``a good job with the image grain''} \\
        & \textit{``excellent use of image grain''} \\
    \midrule
        \multirow{5}{*}{Light\_On\_White}
        & \textit{``light on white''} \\
        & \textit{``great for the light on white''} \\
        & \textit{``nice light on white''} \\
        & \textit{``love the light on white''} \\
        & \textit{``like the light on white''} \\
    \midrule
        \multirow{5}{*}{Long\_Exposure}
        & \textit{``long exposure''} \\
        & \textit{``nice long exposure''} \\
        & \textit{``nice use of long exposure''} \\
        & \textit{``enjoy these long exposure shots''} \\
        & \textit{``look great with the long exposure''} \\
    \midrule
        \multirow{5}{*}{Macro}
        & \textit{``macro''} \\
        & \textit{``excellent detailed macro''} \\
        & \textit{``nice macro''} \\
        & \textit{``good macro shot''} \\
        & \textit{``great macro''} \\
    \bottomrule
    \end{tabular}
    \begin{tabular}{cc}
    \toprule
        Style &Prompts \\
    \midrule
        \multirow{5}{*}{Motion\_Blur}
        & \textit{``motion blur''} \\
        & \textit{``nice motion blur''} \\
        & \textit{``great use of the motion blur''} \\
        & \textit{``i love the motion blur''} \\
        & \textit{``cool motion blur''} \\
    \midrule
        \multirow{5}{*}{Negative\_Image}
        & \textit{``negative image looks good''} \\
        & \textit{``love the negative images''} \\
        & \textit{``the negative image is captivating''} \\
        & \textit{``use of the negative image is interesting''} \\
        & \textit{``fan of the negative image''} \\
    \midrule
        \multirow{5}{*}{Rule\_of\_Thirds}
        & \textit{``rule of thirds''} \\
        & \textit{``benefited from the rule of thirds''} \\
        & \textit{``followed the rule of thirds nicely''} \\
        & \textit{``use of the rule of thirds is fantastic''} \\
        & \textit{``great use of rule of thirds''} \\
    \midrule
        \multirow{5}{*}{Shallow\_DOF}
        & \textit{``shallow dof''} \\
        & \textit{``nice shallow dof''} \\
        & \textit{``i love the shallow DOF''} \\
        & \textit{``lovely use of shallow DOF''} \\
        & \textit{``shallow dof works perfect here''} \\
    \midrule
        \multirow{5}{*}{Silhouettes}
        & \textit{``silhouettes''} \\
        & \textit{``like the silhouettes''} \\
        & \textit{``great silhouettes''} \\
        & \textit{``i really like silhouettes''} \\
        & \textit{``silhouettes are lovely''} \\
    \midrule
        \multirow{5}{*}{Soft\_Focus}
        & \textit{``soft focus''} \\
        & \textit{``love the soft focus''} \\
        & \textit{``love the effect of soft focus''} \\
        & \textit{``excellent use of soft focus''} \\
        & \textit{``lovely soft focus''} \\
    \midrule
        \multirow{5}{*}{Vanishing\_Point}
        & \textit{``vanishing point''} \\
        & \textit{``i like the lines and fading or vanishing''} \\
        & \textit{``i love the lines and vanishing point''} \\
        & \textit{``nice to see the vanishing point off of center''} \\
        & \textit{``the background with the vanishing point is nice''} \\
    \bottomrule
    \end{tabular}
    \vspace{-1mm}
    \caption{Text prompts used in the ensemble approach for AVA-Style ZSL.}
    \label{tab:style_prompts}
    \vspace{-4mm}
\end{table*}

%% file: Tables/supp-4-iaa-prompts.tex
\begin{table}[!htp]
    \centering
    \footnotesize
    \setlength\tabcolsep{1.2pt}
    \begin{tabular}{lccc}\toprule
    &\multicolumn{2}{c}{Prompts} \\\cmidrule{2-3}
        &$\vect{p}_g$ &$\vect{p}_b$ \\\midrule
    Single Prompt
        &\textit{``good image"} &\textit{``bad image"} \\\midrule
    \multirow{6}{*}{Ensemble of Prompts}
        &\textit{``good image"} &\textit{``bad image"} \\
        &\textit{``good lighting"} &\textit{``bad lighting"} \\
        &\textit{``good content"} &\textit{``bad content"} \\
        &\textit{``good background"} &\textit{``bad background"} \\
        &\textit{``good foreground"} &\textit{``bad foreground"} \\
        &\textit{``good composition"} &\textit{``bad composition"} \\
    \bottomrule
    \end{tabular}
    \vspace{-1mm}
    \caption{Text prompts used in ZSL for IAA.}
    \label{tab:iaa_prompts}
    \vspace{-1mm}
\end{table}

%% file: Tables/supp-7-koniq10k.tex
\begin{table}
\begin{center}
\footnotesize
\begin{tabular}{lccc}\toprule
Method &SRCC &PLCC \\\midrule
BRISQUE \cite{mittal2012no} &0.665 &0.681 \\
ILNIQE \cite{zhang2015feature} &0.507 &0.523 \\
HOSA \cite{xu2016blind} &0.671 &0.694 \\
BIECON \cite{kim2016fully} &0.618 &0.651 \\
WaDIQaM \cite{bosse2017deep} &0.797 &0.805 \\
PQR \cite{zeng2017probabilistic} &0.880 &0.884 \\
SFA \cite{li2018has} &0.856 &0.872 \\
DBCNN \cite{zhang2018blind} &0.875 &0.884 \\
MetaIQA \cite{zhu2020metaiqa}  &0.850 &0.887 \\
BIQA \cite{su2020blindly} &0.906 &0.917 \\
CLIP-IQA$^+$ \cite{wang2022exploring} &0.895 &0.909 \\ 
KonCept512 \cite{koniq10k} &0.921 &0.937 \\
MUSIQ \cite{Ke_2021_ICCV} &0.924 &0.937 \\\midrule
\ours{}-R &0.919 &0.932 \\
\bottomrule
\end{tabular}
\end{center}
\vspace{-2mm}
\caption{Results on KonIQ-10k~\cite{koniq10k} dataset.  We take numbers from \cite{wang2022exploring, Ke_2021_ICCV} for results of the reference methods.} \label{tab:koniq-results}
\vspace{-3mm}
\end{table}

%% file: Tables/supp-8-image-attribution.tex
\begin{table}
\begin{center}
\tiny
\setlength\tabcolsep{1.2pt} 
\begin{tabular}{l|c|c}\toprule
Flickr Link &User &License \\\midrule
\textbf{Figure 3 (from left to right, top to bottom)} & & \\
http://www.flickr.com/photos/43437767@N00/7499578096/ &43437767@N00 &CC BY-SA 2.0 \\
http://www.flickr.com/photos/12708857@N00/228617373/ &12708857@N00 &CC BY-SA 2.0 \\
http://www.flickr.com/photos/39443895202@N01/4295525241/ &39443895202@N01 &CC BY-NC 2.0 \\
http://www.flickr.com/photos/43343993@N00/6814873580/ &43343993@N00 &CC BY-NC-SA 2.0 \\
http://www.flickr.com/photos/93656595@N00/5212354067/ &93656595@N00 &CC BY-NC 2.0 \\
http://www.flickr.com/photos/19761391@N06/6190643783/ &19761391@N06 &CC BY-NC-SA 2.0 \\
http://www.flickr.com/photos/8490344@N04/5805954950/ &8490344@N04 &CC BY-NC-SA 2.0 \\
http://www.flickr.com/photos/28577026@N02/4732322374/ &28577026@N02 &CC BY 2.0 \\
http://www.flickr.com/photos/40595948@N00/4125943270/ &40595948@N00 &CC BY 2.0 \\
http://www.flickr.com/photos/8397802@N05/6782627736/ &8397802@N05 &CC BY-NC-SA 2.0 \\\midrule
\textbf{Figure 4 (from left to right, top to bottom)} & & \\
http://www.flickr.com/photos/28081633@N00/3388712525/ &28081633@N00 &CC BY-SA 2.0 \\
http://www.flickr.com/photos/69078621@N00/2501256504/ &69078621@N00 &CC BY-NC 2.0 \\
http://www.flickr.com/photos/21657526@N00/8460154333/ &21657526@N00 &CC BY-NC-SA 2.0 \\
http://www.flickr.com/photos/31990116@N03/8746457937/ &31990116@N03 &CC BY-SA 2.0 \\
http://www.flickr.com/photos/61585804@N00/4639050491/ &61585804@N00 &CC BY-NC-SA 2.0 \\
http://www.flickr.com/photos/78135748@N00/3824485636/ &78135748@N00 &CC BY 2.0 \\
http://www.flickr.com/photos/86381710@N00/163977327/ &86381710@N00 &CC BY-NC-SA 2.0 \\
http://www.flickr.com/photos/11152520@N03/5700224418/ &11152520@N03 &CC BY 2.0 \\
http://www.flickr.com/photos/77175355@N07/10862487886/ &77175355@N07 &CC BY-NC 2.0 \\
http://www.flickr.com/photos/76042652@N00/9467289840/ &76042652@N00 &CC BY-NC-SA 2.0 \\
http://www.flickr.com/photos/74167788@N00/2746983219/ &74167788@N00 &CC BY-NC 2.0 \\
http://www.flickr.com/photos/16409072@N08/7107776883/ &16409072@N08 &CC BY-NC-SA 2.0 \\
http://www.flickr.com/photos/13447407@N00/205662428/ &13447407@N00 &CC BY-NC 2.0 \\
http://www.flickr.com/photos/60635600@N08/6070993297/ &60635600@N08 &CC BY-NC 2.0 \\
http://www.flickr.com/photos/30626457@N00/9629668489/ &30626457@N00 &CC BY 2.0 \\
http://www.flickr.com/photos/31916492@N02/9702908989/ &31916492@N02 &CC BY-NC 2.0 \\
http://www.flickr.com/photos/24742305@N00/3561351919/ &24742305@N00 &CC BY 2.0 \\
http://www.flickr.com/photos/19072679@N00/8216243918/ &19072679@N00 &CC BY-NC-SA 2.0 \\
http://www.flickr.com/photos/50795598@N02/8532549490/ &50795598@N02 &CC BY-NC-SA 2.0 \\
http://www.flickr.com/photos/40573754@N04/4162338388/ &40573754@N04 &CC BY-NC-SA 2.0 \\\midrule
\textbf{Figure 6 (from left to right, top to bottom)} & & \\
http://www.flickr.com/photos/33602849@N00/1348094685/ &33602849@N00 &CC BY-NC-SA 2.0 \\
http://www.flickr.com/photos/32842313@N00/4435718106/ &32842313@N00 &CC BY-NC-SA 2.0 \\
http://www.flickr.com/photos/37306288@N02/6788530155/ &37306288@N02 &CC BY 2.0 \\
http://www.flickr.com/photos/32535586@N07/8120735257/ &32535586@N07 &CC BY-NC 2.0 \\
http://www.flickr.com/photos/62528187@N00/8596361580/ &62528187@N00 &CC BY 2.0 \\
http://www.flickr.com/photos/36543005@N00/242395156/ &36543005@N00 &CC BY 2.0 \\
http://www.flickr.com/photos/47100034@N08/8968242975/ &47100034@N08 &CC BY-NC-SA 2.0 \\
http://www.flickr.com/photos/8397802@N05/6856483689/ &8397802@N05 &CC BY-NC-SA 2.0 \\
http://www.flickr.com/photos/40355539@N00/4948021193/ &40355539@N00 &CC BY-NC-SA 2.0 \\
http://www.flickr.com/photos/10734170@N08/4049047636/ &10734170@N08 &CC BY-NC-SA 2.0 \\
http://www.flickr.com/photos/83670786@N03/8184338593/ &83670786@N03 &CC BY-NC-SA 2.0 \\
http://www.flickr.com/photos/24328811@N00/5091406011/ &24328811@N00 &CC BY-NC-SA 2.0 \\
http://www.flickr.com/photos/46318514@N06/4911877298/ &46318514@N06 &CC BY-NC-SA 2.0 \\
http://www.flickr.com/photos/16482030@N00/5593982816/ &16482030@N00 &CC BY-NC 2.0 \\
http://www.flickr.com/photos/68683191@N00/7915207374/ &68683191@N00 &CC BY-SA 2.0 \\
http://www.flickr.com/photos/51963363@N00/5921291430/ &51963363@N00 &CC BY-NC-SA 2.0 \\
http://www.flickr.com/photos/52713160@N00/6422846525/ &52713160@N00 &CC BY-NC-SA 2.0 \\
http://www.flickr.com/photos/28658116@N02/7914158128/ &28658116@N02 &CC BY-SA 2.0 \\
http://www.flickr.com/photos/10957255@N08/9365989820/ &10957255@N08 &CC BY-NC-SA 2.0 \\
http://www.flickr.com/photos/38439215@N06/4809080599/ &38439215@N06 &CC BY-NC-SA 2.0 \\
http://www.flickr.com/photos/92755733@N00/3017896883/ &92755733@N00 &CC BY 2.0 \\
http://www.flickr.com/photos/33455872@N05/7443887988/ &33455872@N05 &CC BY-NC-SA 2.0 \\
http://www.flickr.com/photos/8833673@N05/4462477090/ &8833673@N05 &CC BY-NC-SA 2.0 \\
http://www.flickr.com/photos/54852753@N05/5766806996/ &54852753@N05 &CC BY 2.0 \\
http://www.flickr.com/photos/90088957@N00/6434876963/ &90088957@N00 &CC BY-NC 2.0 \\
\bottomrule
\end{tabular}
\end{center}
\vspace{-2mm}
\caption{Flickr links to the KonIQ-10k~\cite{koniq10k} images shown in the paper. } \label{tab:koniq-attribution}
\vspace{-3mm}
\end{table}